\documentclass[letterpaper, 10 pt, conference]{ieeeconf}  

\IEEEoverridecommandlockouts                              

\usepackage{cite}
\usepackage{amsmath,amssymb,amsfonts}
\usepackage{algorithmic}
\usepackage{graphicx}
\usepackage{textcomp}
\usepackage{xcolor}
\usepackage{etoolbox}
\usepackage{graphicx}
\usepackage[skip=0.333\baselineskip]{caption}
\usepackage{subcaption}
\usepackage{multirow}
\usepackage{multicol}
\usepackage{hyperref}
\usepackage{graphicx}
\usepackage{amsmath}
\usepackage{amssymb}
\usepackage{booktabs}
\usepackage{color, xcolor}
\usepackage{multirow}
\usepackage{makecell}

\usepackage{cuted}

\title{\LARGE \bf
METDrive: Multimodal End-to-end Autonomous Driving with Temporal Guidance
}

\author{
    Ziang Guo$^{1}$,  
    Xinhao Lin$^{2}$, 
    Zakhar Yagudin$^{1}$, 
    Artem Lykov$^{1}$, \\
    Yong Wang$^{2}$, 
    Yanqiang Li$^{2}$ and 
    Dzmitry Tsetserukou$^{1}$
    \thanks{$^{1}$Ziang Guo, Zakhar Yagudin, Artem Lykov, Dzmitry Tsetserukou are with the Intelligent Space Robotics Laboratory, Center for Digital Engineering, Skolkovo Institute of Science and Technology, Moscow, Russia.
    \tt \{ziang.guo, Zakhar.Yagudin, Artem.Lykov, d.tsetserukou\}@skoltech.ru}
    \thanks{$^{2}$Xinhao Lin, Yong Wang, Yanqiang Li are with the Institute of Automation, Qilu University of Technology (Shandong Academy of Sciences), Jinan, 250399, P.R.China.
    \tt cxlxh727@gmail.com, \{liyq, wangyong1\}@sdas.org}
    }

\begin{document}

\maketitle
\thispagestyle{empty}
\pagestyle{empty}

\begin{abstract}

Multimodal end-to-end autonomous driving has shown promising advancements in recent work. By embedding more modalities into end-to-end networks, the system's understanding of both static and dynamic aspects of the driving environment is enhanced, thereby improving the safety of autonomous driving. In this paper, we introduce METDrive, an end-to-end system that leverages temporal guidance from the embedded time series features of ego states, including rotation angles, steering, throttle signals, and waypoint vectors. The geometric features derived from the perception sensor data and the time series features of ego state data jointly guide the waypoint prediction with the proposed temporal guidance loss function. We evaluated METDrive on the CARLA leaderboard benchmarks, achieving a driving score of $70\%$, a route completion score of $94\%$, and an infraction score of $0.78$.

\end{abstract}

\section{Introduction}

Multimodal end-to-end autonomous driving systems have shown significant promise in enhancing the robustness and reliability of autonomous vehicles \cite{chen2024end}, \cite{omeiza2021explanations}. Sensor fusion, which integrates data from multiple sources such as cameras and LiDAR, has become a cornerstone in advancing these systems \cite{singh2023fusion}, \cite{choi2023semanticsfusion}. However, the raw data acquired from the perception sensors frequently include a multitude of irrelevant objects that do not significantly contribute to the ego vehicle's motion planning. Consequently, the output of an end-to-end model that relies on the fused features of these sensors may be adversely affected by the presence of such irrelevant detection\cite{wang2022towardsfusion}, \cite{malawade2022hydrafusion}. We show that existing end-to-end methodologies, which do not incorporate ego-related states, exhibit shortcomings and failure cases in certain scenarios. In this paper, we propose that the inclusion of additional ego-related features as guidance for the encoders can serve as a viable solution to mitigate this issue. Thus, we present METDrive, a novel end-to-end system that incorporates temporal guidance based on inputs from ego states.

\begin{figure}
    \centering
    \includegraphics[width=0.95\linewidth]{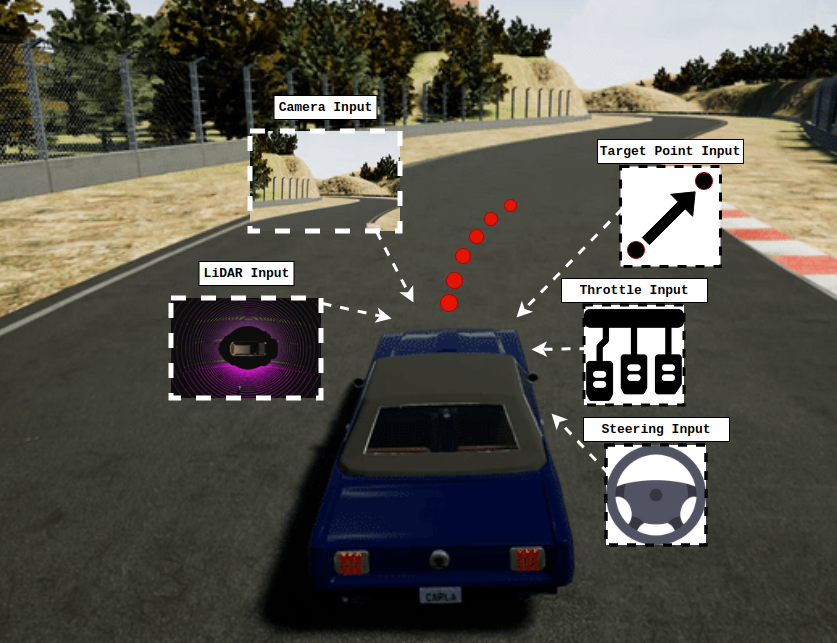}
    \caption{We propose encoding additional modalities from the ego vehicle's states to guide planning tasks in end-to-end autonomous driving systems. Specifically, we treat sequences of steering, throttle, rotation angle, and waypoint vector inputs as time series data. The fused geometric and time series features then guide the waypoint prediction in our METDrive.}
    \label{fig:MET_feature}
\vspace{-0.5cm}
\end{figure}

\par Furthermore, critical ego-related states, such as rotation angles, steering inputs, throttle levels, and waypoints, are typically represented in different modalities compared to the raw images and the LiDAR point clouds. To extend the capabilities of the end-to-end autonomous driving system, we propose a novel approach that encodes these ego-related states as time series features. By incorporating these temporal cues as illustrated in Fig. \ref{fig:MET_feature}, we aim to guide the geometric features extracted from the perception sensors, thereby improving the system's ability to make informed decisions. This integration not only enhances the system's understanding of its environment, but also facilitates more accurate and context-aware motion planning. \par The CARLA leaderboard \cite{carlaLeaderboard} offers a challenging online evaluation platform for autonomous agents operating within the CARLA simulator, including extended routes designed to test the endurance and reliability of autonomous systems. To evaluate the performance of our pipeline on long-term tasks, we selected the route completion, infraction score, and driving score benchmark for evaluation. \par Based on the above insights, our contributions are concluded as follows:
\begin{itemize}
    \item We observed that recent end-to-end models do not incorporate ego states as input, which is essential for motion planning tasks. We show that these models may encounter failure cases in simple scenarios.
    \item We propose a time series data encoder for processing the ego-related states including rotation angle, steer signal, throttle signal, and waypoint vectors to guide the fusion of features from all the sensors. Consequently, a temporal guidance loss is designed to optimize the consistency between the predictions of waypoints from adjacent time steps. Based on such a design, METDrive is presented as a novel end-to-end pipeline with temporal ego state's guidance.
    \item In the CARLA leaderboard benchmarks, our proposed pipeline achieved a better driving score, route completion, and infraction score compared to other recent approaches.
\end{itemize}

\begin{figure}
    \centering
    \includegraphics[width=0.9\linewidth]{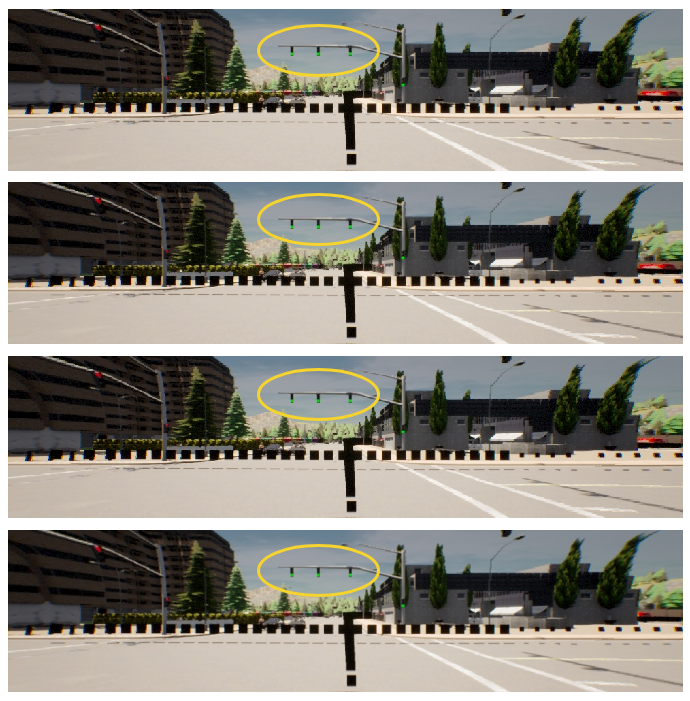}
    \caption{During the experiments, it was observed that the ego vehicle halted at the crossing despite the presence of a green traffic light marked in yellow circles and clear traffic conditions. During these consecutive frames, the ego vehicle remained stationary and did not proceed along the predefined ground truth waypoints that are marked in black dots on the map.}
    \label{fig:failure}
\vspace{-0.3cm}
\end{figure}

\section{Related Work}

In the field of autonomous driving, particularly in end-to-end learning systems, significant strides have been made in leveraging deep learning architectures to improve vehicle navigation and decision-making processes \cite{zhou2024embodied}, \cite{guo2024co}, \cite{gbagbe2024bi}, \cite{guo2025vdt}. The following section reviews recent advances in model architectures for self-driving cars, focusing on their contributions and relevance to their proposed models.

\subsection{Learning from All Vehicles}
A key advancement in motion planning within end-to-end systems is the concept of learning from all vehicles in the environment, as proposed by Chen et al. \cite{chen2022learning}. Their model, which integrates a perception module, a motion planner, and a low-level controller, utilizes a three-stage modular pipeline. This approach significantly improves the generalization of motion planning by training on the trajectories of all surrounding vehicles, rather than just the ego vehicle. The model architecture is designed to produce vehicle-invariant features that enhance the motion planner's ability to predict future trajectories across different vehicles.

\subsection{Trajectory Prediction and Multimodal Fusion}
The trajectory prediction has been a central focus in recent models, where the objective is to generate accurate waypoints for the future path of the vehicle. The TransFuser model, introduced by Chitta et al. \cite{chitta2022transfuser}, employs a multimodal fusion transformer that combines RGB images and LiDAR data. This architecture allows the model to leverage the complementary nature of different sensor inputs, thus improving the accuracy of waypoint predictions. The model's use of self-attention mechanisms within the transformer architecture is particularly effective in incorporating global context into the decision-making process. 

\subsection{Addressing Biases in Imitation Learning}
A significant challenge in imitation learning-based autonomous driving models is the presence of hidden biases, particularly those related to lateral recovery and longitudinal control. Jaeger et al. \cite{jaeger2023hidden} addressed these issues by identifying biases in state-of-the-art models that rely on the following of the target point and the prediction of multimodal waypoints. Their proposed model, TransFuser++, mitigates these biases through a combination of architectural modifications and training strategies, resulting in improved driving performance on benchmark tasks. 

\subsection{Combining Trajectory Planning with Control Prediction}
Recent approaches have also explored the integration of trajectory planning with control prediction. For instance, a novel architecture proposed by Wu et al. \cite{wu2022trajectory} combines these two paradigms within a single learning pipeline. The model uses a multi-step control prediction branch guided by a trajectory planning branch, which allows for more accurate and context-aware control decisions. This approach is particularly beneficial for handling complex driving scenarios, where direct prediction of control actions can lead to suboptimal behavior.

\subsection{Alignment with Student's Perception and Teacher's Planning}
Jia et al. \cite{jia2023driveadapter} introduced a novel approach to end-to-end autonomous driving using a frozen teacher model for planning while the student model focuses on perception. The paper proposes an adapter module to align the student's perception output with the teacher's planning input, addressing the distribution gap between predicted and ground-truth data incorporating action-guided feature learning and a masking strategy to refine the learning process.

\section{Temporal Guidance for End-to-end Autonomous Driving}

\textbf{Imitation Learning with More Modalities.} Through the experiments we conducted on the recent end-to-end systems in CARLA \cite{chitta2022transfuser}, \cite{jaeger2023hidden}, where we evaluated the provided checkpoints from their papers in the long routes of CARLA towns. Regarding the performance of TransFuser++ \cite{jaeger2023hidden}, Fig. \ref{fig:failure} from the experiments shows that the ego vehicle unexpectedly stopped at a green light, even when traffic conditions were clear. Based on more observation, we found that, in some cases, the agent's behavior could encounter failure even when the scenarios complied with the distribution of training data. To address these shortcomings, our proposed system aims to encode more modalities, such as ego vehicle's states, to provide enhanced guidance for motion planning tasks. Considering the above failure cases, we introduce the ego states to our training data to enable our model with additional and critical imitation for motion output. In the following, we demonstrate the details of our approach and the experimental results.

\begin{figure*}[t]
    \centering
    \includegraphics[width=0.9\linewidth]{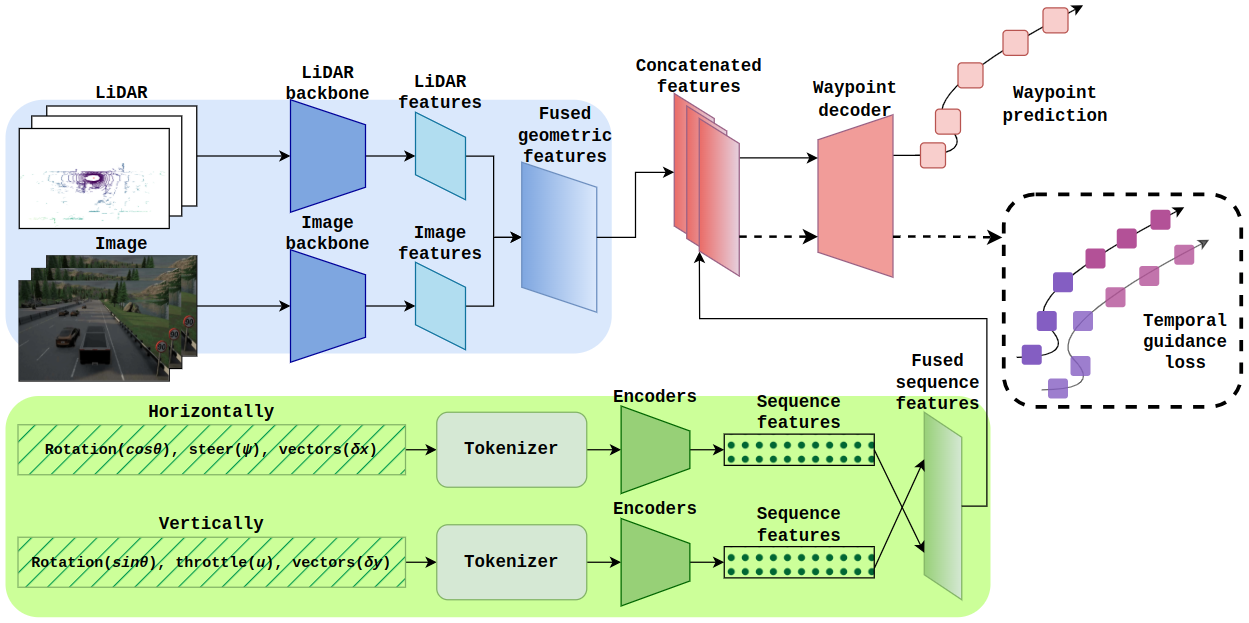}
    \caption{Our METDrive comprises a perception branch and a temporal branch. In the perception branch, features of the image and LiDAR point clouds are fused as geometric features. In the temporal branch, sequences of ego state inputs are encoded as time series features. The decomposed time series features are integrated through a fully connected network by learning the mapping that optimally captures their interactions with trainable weights. The geometric and time series features are then concatenated and fed into the waypoint decoders. To guide this process, importance-sampled concatenated features based on close and far temporal proximity additionally generate two sets of waypoint outputs through waypoint decoders, and the temporal guidance loss function minimizes the L2 norm between the corresponding parts of the predicted trajectory from different temporal proximity.}
    \label{fig:MET}
\vspace{-0.3cm}
\end{figure*}

\textbf{Ego States as Temporal Information.} In contrast to other current end-to-end systems, our proposed system additionally incorporates ego-related data input, thereby enhancing the system's ability to leverage temporal information \cite{zhang2024ego}. Specifically, we treat ego-related states as time series data, enabling the system to capture the dynamic evolution of the ego vehicle's state over time \cite{greer2024egoperception}. Consequently, our system architecture is divided into two complementary branches: a perception branch, which processes sensor data from cameras and LiDAR, transforming these inputs into geometric features that represent the spatial layout of the environment; and a temporal branch, which processes the ego-related data as time-series inputs, thereby capturing the temporal dynamics of the ego vehicle's state. In Fig. \ref{fig:MET}, we illustrate our proposed approach, in which geometric and temporal characteristics are first extracted by their respective encoders. Subsequently, these encoded features are fused and processed through Gated Recurrent Units (GRUs) \cite{cho2014gru} to facilitate waypoint prediction with temporal guidance. \par \textbf{Perception Branch.} The perception branch includes image and LiDAR encoders, both based on ResNet, which extract geometric features from sensor data \cite{he2016resnet}. An attention-based feature fusion block is used to integrate these features effectively \cite{guo2024hawkdrive}. Through the use of attention-based fusion, geometric features are processed in conjunction with temporal input from both image and LiDAR data, thereby ensuring consistency for subsequent fusion with time series data \cite{renz2022plant}.
\par \textbf{Temporal Branch.} Complementing the temporal perception inputs, the ego vehicle's rotation angles $\theta$, steering $\psi$, throttle $u$ signals and the normalized vectors ($\delta x, \delta y$) between target points are encoded accordingly. To clarify the lateral and longitudinal features for obtaining single-dimensional time series information, these temporal signals are decomposed into horizontal and vertical directions within the ego vehicle coordinate system for tokenization. To obtain the embedded horizontal tokens $T_x \in \mathbb{R}^{1 \times 3B \times B}$, where $B$ is the batch size, the input sequence data are processed via positional and token embedding, as follows:

\begin{equation}
\begin{aligned}
    T_x &= [T_{x1}, T_{x2}, T_{x3}], \\
    T_{x1} &= E_{pos}(\cos{\theta}) \oplus E_{token}(\cos{\theta}), \\
    T_{x2} &= E_{pos}(\psi) \oplus E_{token}(\psi), \\
    T_{x3} &= E_{pos}(\delta x) \oplus E_{token}(\delta x).
\end{aligned}
\end{equation}
where $E_{pos}$ denotes the positional embedding, ensuring unique positions of the elements in the series for the encoders with the index of the dimension $i$ and the dimensions of embedding. It is defined by the following equation: 
\begin{equation}
PE(position, 2i) = \sin\left(\frac{position}{10000^{2i / dimension}}\right),
\end{equation}

\begin{equation}
PE(position, 2i+1) = \cos\left(\frac{position}{10000^{2i / dimension}}\right).
\end{equation}
$E_{token}$ denotes the token embedding, extracting the features of the input sequence data via a 1D convolution layer. \par Similarly, the embedded vertical tokens $T_y \in \mathbb{R}^{1 \times 3B \times B}$ is computed with $\sin{\theta}, u$, and $\delta y$.
\begin{equation}
\begin{aligned}
    T_y &= [T_{y1}, T_{y2}, T_{y3}], \\
    T_{y1} &= E_{pos}(\sin{\theta}) \oplus E_{token}(\sin{\theta}), \\
    T_{y2} &= E_{pos}(u) \oplus E_{token}(u), \\
    T_{y3} &= E_{pos}(\delta y) \oplus E_{token}(\delta y).
\end{aligned}
\end{equation}
\par The corresponding tokens are subsequently encoded using self-attention-based encoders. The encoded features are then integrated via a fully connected network, which produces the fused features from the temporal inputs of rotation angles, steering angles, throttle signals, and the normalized vectors of target points. \par The temporal geometric and time series features are concatenated and fed into a GRU-based waypoint decoder. Along with the target point input, this setup enables the regressive prediction of the output waypoints as in Transfuser \cite{chitta2022transfuser}. 
\par \textbf{Temporal Guidance Loss.} To ensure guidance based on the sequence input of ego states, a loss function is designed to minimize the differences between two predictions of waypoints from different temporal proximity. This is achieved through importance sampling of the fused features derived from geometric and time series features, where the fused features from the perception and states closer to the current time are assigned higher importance. The temporal guidance loss $\mathcal{L}_\mathrm{Temporal}$ is denoted as follows:

\begin{equation}
\begin{split}
\mathcal{L}_\mathrm{Temporal} = \alpha (\left\| \mathbf{y}(\mathbf{[F_g, F_t]}_{t=0}^{t=\frac{n}{2}}) - \mathbf{\hat{y}}(\mathbf{[F_g, F_t]})_{[0, ..., \frac{n}{2}]} \right\|^2_2) \\
+ \beta(\left\| \mathbf{y}(\mathbf{[F_g, F_t]}_{t=\frac{n}{2}}^{t=n}) - \mathbf{\hat{y}}(\mathbf{[F_g, F_t]})_{[\frac{n}{2}, ..., n]} \right\|^2_2 ),
\end{split}
\end{equation}
where $\mathbf{F_g, F_t}$ are the geometric and time series features from the encoders and $\alpha$, $\beta$ are the importance weights. In $n$ time steps, the fused features are masked for the time steps $0$ to $\frac{n}{2}$ and the time steps $\frac{n}{2}$ to $n$, and then passed to the waypoint decoder to obtain partial predicted trajectory separately. The features of the $n$ time steps are decoded as full trajectory prediction from an entire temporal span. The temporal guidance loss is to minimize the L2 norm of the corresponding parts of the predicted trajectory.

\section{Experimental Results}

\begin{table}[t]
\vspace{0.13cm}
\centering
\caption{Performance comparison on CARLA Longest6 benchmark.}
\resizebox{0.45\textwidth}{!}{
\begin{tabular}{c|c c c}
\toprule
     Method & DS $\uparrow$ & RC $\uparrow$ & IS $\uparrow$ \\ \hline
    \textbf{Ours} & $\textbf{70}$ & $\textbf{94}$ & $0.78$  \\
    TransFuser++ \cite{jaeger2023hidden} & $69$ & $94$ & $0.72$  \\
    ThinkTwice \cite{jia2023thinktwice} & $67$ & $77$ & $\textbf{0.84}$  \\
    DriveAdapter \cite{jia2023driveadapter} & $59$ & $82$ & $0.68$  \\
    CaT \cite{zhang2023coaching} & $58$ & $79$ & $0.77$  \\
    InteractionNet \cite{fu2023interactionnet} & $51$ & $87$ & $0.60$  \\
    TCP \cite{wu2022trajectory} & $48$ & $72$ & $0.65$  \\
    \bottomrule
\end{tabular}}
\label{table:leaderboard}
\vspace{-0.3cm}
\end{table}

\textbf{Experimental Setup.} The dataset for our training is collected in the CARLA Towns 01, 03, 04, 06, 07, and 10 with front camera images, central LiDAR point clouds, rotation angle record, control signal record, velocity record, control command record, and target point record. As the recordings are based on the autonomous agents provided in CARLA, whose driving behaviors are not as natural and smooth as those of human drivers, we filtered out and averaged the noisy sequence from control signals to minimize the degradation of our model's performance. \par In \cite{jaeger2023hidden}, the training receipt contains a two-stage training, where the perception branch is trained with the corresponding losses first. Then a fine-tuning is performed on the checkpoint with all the losses. We trained METDrive in one stage with both the perception branch and the temporal branch. With 320K training samples, we trained our model on a single Nvidia RTX 4090 24G at batch size 16.

\begin{figure*}
    \centering
    \includegraphics[width=1\linewidth]{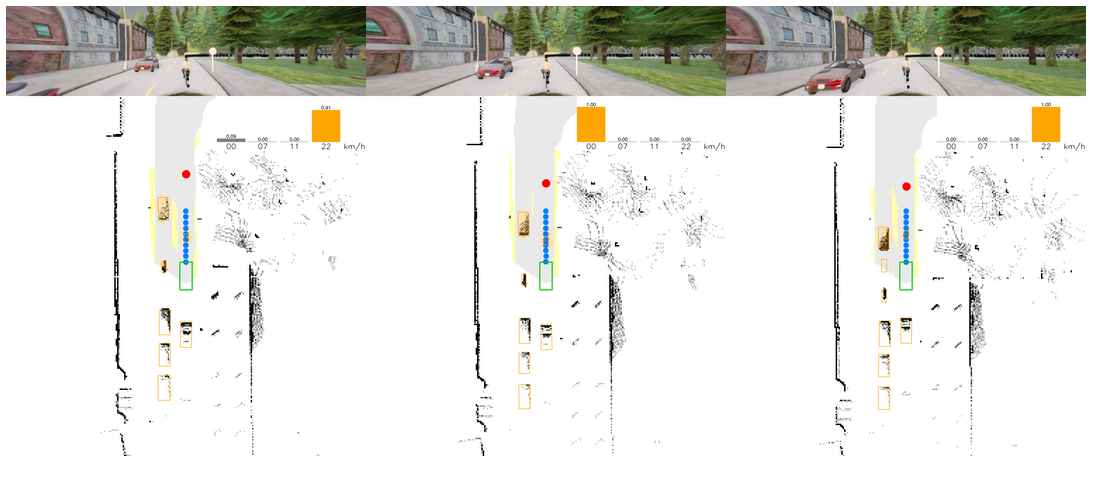}
    \caption{In the experiments of CARLA Town 02, the baseline model's speed output oscillated between $0$ km/h and $22$ km/h during the continuous frames resulting in the jerks in driving.}
    \label{fig:baseline}
\end{figure*}

\begin{figure*}
    \centering
    \includegraphics[width=1\linewidth]{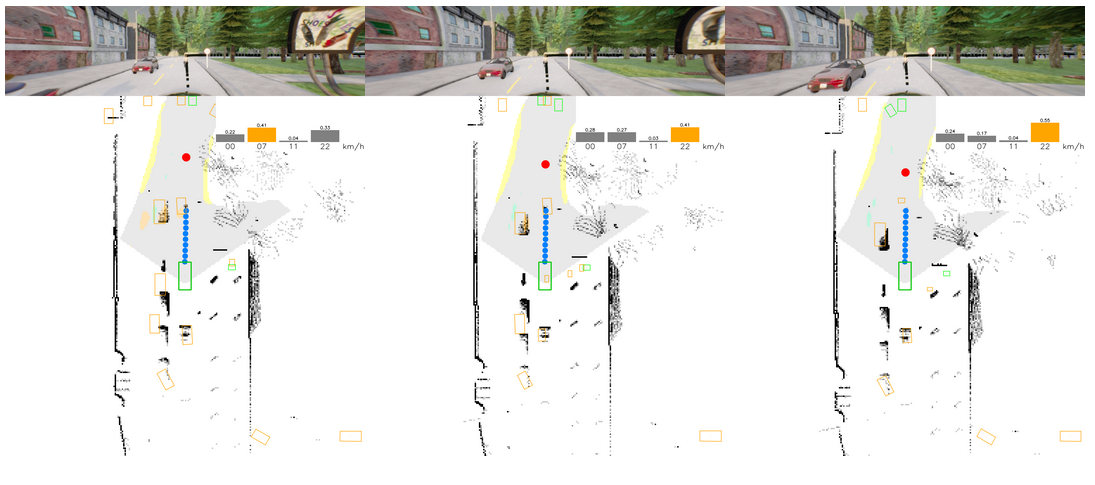}
    \caption{In a similar scenario as above, our METDrive provided natural driving behavior with smooth speed output ranging from $7$ km/h to $22$ km/h without sudden jerks.}
    \label{fig:viz}
\vspace{-0.3cm}
\end{figure*}

\par \textbf{Experimental Results.} To evaluate the performance of our model, the CARLA Longest6 benchmarks are used \cite{carlaLeaderboard}. Table \ref{table:leaderboard} shows the performance comparison of the models in the evaluation, where our proposed system achieves a better driving score, route completion, and infraction score compared to the reproduced results from the recent approaches. In Table \ref{table:leaderboard}, RC means the Route Completion, which denotes the percentage of the route distance completed by an agent. IS means the Infraction Score that aggregates all tracked infractions triggered by an agent as a geometric series. DS means Driving Score, which is the product between the route completion and the infraction penalty. \par In Fig. \ref{fig:baseline} and Fig. \ref{fig:viz}, we intend to highlight the effectiveness of integrating the ego states with geometric features. The visualization from TransFuser++ \cite{jaeger2023hidden} provides information such as the front camera input, the model waypoint output in blue points, speed output in a bar chart, etc. We selected a period of continuous frames from the same scenario in CARLA Town 02 for comparison, where Fig. \ref{fig:baseline} is based on the evaluation of TransFuser++ and Fig. \ref{fig:viz} is from our METDrive. The speed output generated by TransFuser++ exhibited unnatural and discrete characteristics, oscillating between extreme values of $0$ km/h and $22$ km/h. This abrupt change in speed resulted in significant driving jerks. In contrast, our proposed METDrive generated speed outputs that avoid such extreme oscillations, thereby ensuring continuous and smooth driving paths.

\begin{table}[]
     \centering
    \caption{The ablation experiment by training without temporal guidance loss.}
    \resizebox{0.45\textwidth}{!}{
    \begin{tabular}{c|ccc}
        Method & DS $\uparrow$ & RC $\uparrow$ & IS $\uparrow$ \\ \hline
        w/o temporal guidance loss & 68 & 89 & 0.70
    \end{tabular}}
    \label{tab:ablation_loss}
\vspace{-0.5cm}
\end{table}

\textbf{Ablation Study.} To verify the effectiveness of our design, we performed ablation experiments with temporal guidance loss and different sets of sequence input for the time series features. The results of the temporal guidance loss are shown in Table \ref{tab:ablation_loss}. Via the training without the temporal guidance loss, the performance of our METDrive on the CARLA Longest6 benchmark declined, as the encoded time series features were unable to align with the temporal geometric features without the constraints imposed by the designed loss function. \par In Table \ref{tab:ablation_input}, we tested different types of sequence input for tokenizers and found that the decomposed sequence input was optimal. This was because the decomposition clarified and aligned the lateral and longitudinal features along the horizontal and vertical axes of the ego coordinate system.

\begin{table}[t]
    \centering
    \caption{The ablation experiment with different types of sequence input for the tokenizers, where we input the below sequences without decomposition towards the tokenizer.}
    \resizebox{0.46\textwidth}{!}{
    \begin{tabular}{c|ccc}
        Method & DS $\uparrow$ & RC $\uparrow$ & IS $\uparrow$ \\ \hline
        $(\theta_1, \theta_2)+(u_1, u_2)+(\psi_1, \psi_2)+\delta$ & 66 & 87 & 0.69 \\
        $\theta+u+\psi$ & 65 & 83 & 0.64
    \end{tabular}}
    \label{tab:ablation_input}
\vspace{-0.3cm}
\end{table}

\section{Conclusion and Future Work}

In this paper, we identify a common shortcoming in the existing end-to-end systems for the CARLA simulator: the absence of guidance from additional modalities, which may lead to failure cases in long-term tasks. To address this issue, we propose to leverage the ego-related temporal cues, such as rotation angles, steering, throttle signals, and waypoint vectors, to guide the geometric features derived from the perception sensors in predicting waypoints. Consequently, we design a temporal guidance loss to monitor this integration process. To evaluate our system, we conducted experiments on the CARLA Longest6 benchmark, achieving the driving score of $70\%$, route completion score of $94\%$, and infraction score of $0.78$. \par Despite time series features are used as temporal guidance in this study, the efficacy of the feature fusion approach is constrained by the quality of the ego state inputs derived from the collected dataset. Specifically, given that the dataset was generated using an autonomous agent within the CARLA simulation environment, the agent's performance was characterized by inconsistencies, resulting in a lack of smoothness and natural driving behavior. Consequently, the ego state data were replete with significant noise, preprocessing steps such as filtering before they could be utilized effectively for our training. \par In light of these limitations, our future research routine will focus on the acquisition of a higher fidelity dataset. This dataset will be sourced from the expert human drivers who navigate a diverse range of traffic scenarios. By incorporating data from the skilled human operators, we anticipate a reduction in the aforementioned noise and a resultant enhancement in the quality of the input data. Additionally, we intend to develop a more robust framework that encompasses a broader spectrum of time series encoders and feature fusion tailored for ego state inputs. This enhanced approach is expected to mitigate current issues and improve the overall performance and reliability of the system under development.

\section*{Acknowledgements} 
Research reported in this publication was financially supported by RSF grant No. 24-41-02039.




\bibliographystyle{IEEEtran}
\bibliography{references}

\begin{thebibliography}{10}
\providecommand{\url}[1]{#1}
\csname url@samestyle\endcsname
\providecommand{\newblock}{\relax}
\providecommand{\bibinfo}[2]{#2}
\providecommand{\BIBentrySTDinterwordspacing}{\spaceskip=0pt\relax}
\providecommand{\BIBentryALTinterwordstretchfactor}{4}
\providecommand{\BIBentryALTinterwordspacing}{\spaceskip=\fontdimen2\font plus
\BIBentryALTinterwordstretchfactor\fontdimen3\font minus \fontdimen4\font\relax}
\providecommand{\BIBforeignlanguage}[2]{{%
\expandafter\ifx\csname l@#1\endcsname\relax
\typeout{** WARNING: IEEEtran.bst: No hyphenation pattern has been}%
\typeout{** loaded for the language `#1'. Using the pattern for}%
\typeout{** the default language instead.}%
\else
\language=\csname l@#1\endcsname
\fi
#2}}
\providecommand{\BIBdecl}{\relax}
\BIBdecl

\bibitem{chen2024end}
L.~Chen, P.~Wu, K.~Chitta, B.~Jaeger, A.~Geiger, and H.~Li, ``End-to-end autonomous driving: Challenges and frontiers,'' \emph{IEEE Transactions on Pattern Analysis and Machine Intelligence}, vol.~46, no.~12, pp. 10\,164--10\,183, 2024.

\bibitem{omeiza2021explanations}
D.~Omeiza, H.~Webb, M.~Jirotka, and L.~Kunze, ``Explanations in autonomous driving: A survey,'' \emph{IEEE Transactions on Intelligent Transportation Systems}, vol.~23, no.~8, pp. 10\,142--10\,162, 2021.

\bibitem{singh2023fusion}
A.~Singh, ``Transformer-based sensor fusion for autonomous driving: A survey,'' in \emph{Proc. IEEE/CVF Int. Conf. on Computer Vision (ICCV)}, 2023, pp. 3312--3317.

\bibitem{choi2023semanticsfusion}
H.-S. Choi, J.~Jeong, Y.~H. Cho, K.-J. Yoon, and J.-H. Kim, ``Semantics-guided transformer-based sensor fusion for improved waypoint prediction,'' \emph{arXiv preprint arXiv:2308.02126}, 2023.

\bibitem{wang2022towardsfusion}
Z.~Wang, X.~Zeng, S.~L. Song, and Y.~Hu, ``Towards efficient architecture and algorithms for sensor fusion,'' \emph{arXiv preprint arXiv:2209.06272}, 2022.

\bibitem{malawade2022hydrafusion}
A.~V. Malawade, T.~Mortlock, and M.~A. Al~Faruque, ``Hydrafusion: Context-aware selective sensor fusion for robust and efficient autonomous vehicle perception,'' in \emph{Proc. ACM/IEEE Int. Conf. on Cyber-Physical Systems (ICCPS)}, 2022, pp. 68--79.

\bibitem{carlaLeaderboard}
\BIBentryALTinterwordspacing
``Carla leaderboard.'' [Online]. Available: \url{https://leaderboard.carla.org/leaderboard/}
\BIBentrySTDinterwordspacing

\bibitem{zhou2024embodied}
Y.~Zhou, L.~Huang, Q.~Bu, J.~Zeng, T.~Li, H.~Qiu, H.~Zhu, M.~Guo, Y.~Qiao, and H.~Li, ``Embodied understanding of driving scenarios,'' \emph{arXiv preprint arXiv:2403.04593}, 2024.

\bibitem{guo2024co}
Z.~Guo, A.~Lykov, Z.~Yagudin, M.~Konenkov, and D.~Tsetserukou, ``Vlm-auto: Vlm-based autonomous driving assistant with human-like behavior and understanding for complex road scenes,'' in \emph{Proc. Int. Conf. on Foundation and Large Language Models (FLLM)}, 2024, pp. 501--507.

\bibitem{gbagbe2024bi}
K.~F. Gbagbe, M.~A. Cabrera, A.~Alabbas, O.~Alyunes, A.~Lykov, and D.~Tsetserukou, ``Bi-vla: Vision-language-action model-based system for bimanual robotic dexterous manipulations,'' in \emph{Proc. Int. Conf. on Systems, Man, and Cybernetics (SMC)}, 2024, pp. 2864--2869.

\bibitem{guo2025vdt}
Z.~Guo, K.~Gubernatorov, S.~Asfaw, Z.~Yagudin, and D.~Tsetserukou, ``Vdt-auto: End-to-end autonomous driving with vlm-guided diffusion transformers,'' \emph{arXiv preprint arXiv:2502.20108}, 2025.

\bibitem{chen2022learning}
D.~Chen, V.~Koltun, and P.~Kr{\"a}henb{\"u}hl, ``Learning from all vehicles: Motion planning and control with vehicle-invariant perception,'' in \emph{Proc. IEEE/CVF Conf. on Computer Vision and Pattern Recognition (CVPR)}, 2022, pp. 11\,763--11\,773.

\bibitem{chitta2022transfuser}
K.~Chitta, A.~Prakash, B.~Jaeger, Z.~Yu, K.~Renz, and A.~Geiger, ``Transfuser: Imitation with transformer-based sensor fusion for autonomous driving,'' \emph{IEEE Transactions on Pattern Analysis and Machine Intelligence}, vol.~45, no.~11, pp. 12\,878--12\,895, 2022.

\bibitem{jaeger2023hidden}
B.~Jaeger, K.~Chitta, and A.~Geiger, ``Hidden biases of end-to-end driving models,'' in \emph{Proc. IEEE/CVF Int. Conf. on Computer Vision (ICCV)}, 2023, pp. 8240--8249.

\bibitem{wu2022trajectory}
P.~Wu, X.~Jia, L.~Chen, J.~Yan, H.~Li, and Y.~Qiao, ``Trajectory-guided control prediction for end-to-end autonomous driving: A simple yet strong baseline,'' \emph{Advances in Neural Information Processing Systems}, vol.~35, pp. 6119--6132, 2022.

\bibitem{jia2023driveadapter}
X.~Jia, Y.~Gao, L.~Chen, J.~Yan, P.~L. Liu, and H.~Li, ``Driveadapter: Breaking the coupling barrier of perception and planning in end-to-end autonomous driving,'' in \emph{Proc. IEEE/CVF Int. Conf. on Computer Vision (ICCV)}, 2023, pp. 7953--7963.

\bibitem{zhang2024ego}
Z.~Zhang, C.~Wang, W.~Zhao, M.~Cao, and J.~Liu, ``Ego vehicle trajectory prediction based on time-feature encoding and physics-intention decoding,'' \emph{IEEE Transactions on Intelligent Transportation Systems}, vol.~25, no.~7, pp. 6527--6542, 2024.

\bibitem{greer2024egoperception}
R.~Greer and M.~Trivedi, ``Perception without vision for trajectory prediction: Ego vehicle dynamics as scene representation for efficient active learning in autonomous driving,'' \emph{arXiv preprint arXiv:2405.09049}, 2024.

\bibitem{cho2014gru}
K.~Cho, ``Learning phrase representations using rnn encoder-decoder for statistical machine translation,'' \emph{arXiv preprint arXiv:1406.1078}, 2014.

\bibitem{he2016resnet}
K.~He, X.~Zhang, S.~Ren, and J.~Sun, ``Deep residual learning for image recognition,'' in \emph{Proc. IEEE Int. Conf. on Computer Vision and Pattern Recognition (CVPR)}, 2016, pp. 770--778.

\bibitem{guo2024hawkdrive}
Z.~Guo, S.~Perminov, M.~Konenkov, and D.~Tsetserukou, ``Hawkdrive: A transformer-driven visual perception system for autonomous driving in night scene,'' in \emph{Proc. IEEE Intelligent Vehicles Symposium (IV)}, 2024, pp. 2598--2603.

\bibitem{renz2022plant}
K.~Renz, K.~Chitta, O.-B. Mercea, A.~Koepke, Z.~Akata, and A.~Geiger, ``Plant: Explainable planning transformers via object-level representations,'' \emph{arXiv preprint arXiv:2210.14222}, 2022.

\bibitem{jia2023thinktwice}
X.~Jia, P.~Wu, L.~Chen, J.~Xie, C.~He, J.~Yan, and H.~Li, ``Think twice before driving: Towards scalable decoders for end-to-end autonomous driving,'' in \emph{Proc. IEEE/CVF Int. Conf. on Computer Vision and Pattern Recognition (CVPR)}, 2023, pp. 21\,983--21\,994.

\bibitem{zhang2023coaching}
J.~Zhang, Z.~Huang, and E.~Ohn-Bar, ``Coaching a teachable student,'' in \emph{Proc. IEEE/CVF Int. Conf. on Computer Vision and Pattern Recognition (CVPR)}, 2023, pp. 7805--7815.

\bibitem{fu2023interactionnet}
J.~Fu, Y.~Shen, Z.~Jian, S.~Chen, J.~Xin, and N.~Zheng, ``Interactionnet: Joint planning and prediction for autonomous driving with transformers,'' in \emph{Proc. IEEE/RSJ Int. Conf. on Intelligent Robots and Systems (IROS)}, 2023, pp. 9332--9339.

\end{thebibliography}

\end{document}